\documentclass[a4paper]{spie}  

 \usepackage{mathtools}
\usepackage{amsmath,amsfonts,amssymb}
\usepackage{comment}

\usepackage{nicefrac}

\usepackage[caption=false,font=footnotesize,labelformat=empty,position=top]{subfig}
\usepackage{tikz}
\usetikzlibrary{positioning,shapes,arrows}
\usepackage{booktabs}
\usepackage[a4paper, left=1.925cm, right=1.925cm, top=2.54cm, bottom=4.94cm]{geometry}

\usepackage{graphicx}
\graphicspath{{graphics/}}

\usepackage{pgfplots}
\usepackage[justification=justified]{caption}
\pgfplotsset{compat=newest}
\usepgfplotslibrary{groupplots}
\usepgfplotslibrary{dateplot}
\usepackage{varioref}
\usepackage[colorlinks=true, allcolors=blue]{hyperref}
\usepackage[capitalize,noabbrev]{cleveref}

\newcommand{\norm}[1]{\ensuremath\left|\left|#1\right|\right|}
\DeclareMathOperator*{\argmin}{arg\,min}

\title{Superaccurate Camera Calibration via Inverse Rendering}

\author[a]{Morten Hannemose}
\author[b]{Jakob Wilm}
\author[a]{Jeppe Revall Frisvad}
\affil[a]{DTU Compute, Technical University of Denmark}
\affil[b]{SDU Robotics, University of Southern Denmark}

\authorinfo{Further author information: (Send correspondence to M.H.)\\M.H. E-mail: mohan@dtu.dk
}

\begin{document} 
\maketitle

\begin{abstract}
The most prevalent routine for camera calibration is based on the detection of well-defined feature points on a purpose-made calibration artifact. These could be checkerboard saddle points, circles, rings or triangles, often printed on a planar structure. The feature points are first detected and then used in a nonlinear optimization to estimate the internal camera parameters.
We propose a new method for camera calibration using the principle of inverse rendering. Instead of relying solely on detected feature points, we use an estimate of the internal parameters and the pose of the calibration object to implicitly render a non-photorealistic equivalent of the optical features. This enables us to compute pixel-wise differences in the image domain without interpolation artifacts. We can then improve our estimate of the internal parameters by minimizing pixel-wise least-squares differences. In this way, our model optimizes a meaningful metric in the image space assuming normally distributed noise characteristic for camera sensors.
We demonstrate using synthetic and real camera images that our method improves the accuracy of estimated camera parameters as compared with current state-of-the-art calibration routines. Our method also estimates these parameters more robustly in the presence of noise and in situations where the number of calibration images is limited. 
\end{abstract}


\section{INTRODUCTION}
Accurate camera calibration is essential for the success of many optical metrology techniques such as pose estimation, white light scanning, depth from defocus, passive and photometric stereo, and more. To obtain sub-pixel accuracy, it can be necessary to use high-order lens distortion models, but this necessitates a large number of observations to properly constrain the model and avoid local minima during optimization. 

A very commonly used camera calibration routine is that of Zhang\cite{zhang2000flexible}. This is based on detection of feature points, an approximate analytic solution and a nonlinear optimization of the reprojection error to estimate the internal parameters, including lens distortion. Oftentimes, checkerboard corners are detected using Harris' corner detector\cite{harris1988combined}, followed by sub-pixel saddle-point detection, such as that of F{\"o}rstner and G{\"u}lch\cite{forstner1987fast}, which is implemented in OpenCV's \texttt{cornerSubPix()} routine.
This standard technique can be improved for example by more robust and precise sub-pixel corner detectors\cite{geiger2012automatic,chen2018ccdn} or use of a pattern different from the prevalent checkerboard\cite{ding2017robust,ha2017deltille}. A different line of work aims at reducing perspective and lens-dependent bias of sub-pixel estimates\cite{lucchese2002polynomials,placht2014rochade}. In the work of Datta\cite{datta2009accurate}, reprojection errors are reduced significantly by iteratively rectifying images to a frontoparallel view and re-estimating saddle points.
Nevertheless, such techniques are still dependent on how accurately and unbiased the corners/features were detected in the first place. Perspective and lens-distortion are then not considered directly, as their parameters are known only after calibration. Instead, the common approach is to try to make the detector mostly invariant to such effects. However, for larger features such as circles, it is questionable whether these can be detected in an unbiased way without prior knowledge of lens parameters. In addition, the distribution of the localization error is unknown and least-squares optimization may not be optimal.

In this paper, instead of relying solely on the sub-pixel accuracy of points in the image, we render an image of the calibration object given the current estimate of calibration parameters and the pose of the object. 
This non-photorealistic rendering of the texture of the calibration object can be compared to the observed image, which lets us compute pixel-wise differences in the image domain without interpolation. Because we are comparing differences in pixel intensities, we can model the errors as normally distributed which closely resembles the noise characteristics usually seen in camera images. 
This process is iterated in an optimization routine so that we are able to directly minimize the squared difference between the observed pixels and our rendered equivalent.

To ensure convergence of the optimization, the error must be differentiable with respect to camera parameters, object pose, and image coordinates. We ensure this by rendering slightly smoothed versions of the calibration object features.

\section{RELATED WORK}

We use a texture for our implicit rendering. This bears some resemblance to the version of texture-based camera calibration\cite{zhang2011camera} where a known pattern is employed. We thus inherit some of the robustness and accuracy benefits that this method earns because it is not relying exclusively on feature extraction. Our optimization strategy is however simpler and more easily applied in practice as compared with their rank minimization problem with nonlinear constraints.

The work by Rehder \emph{et al.}\cite{rehder2017direct} is more closely related to ours. They argue that an initial selection of feature points (like corners) is an inadequate abstraction. As in our work, they use a standard calibration technique for initialization. With this calibration, they implicitly render the calibration target into selected pixels to get a more direct error formulation based on image intensities. This is then used to further refine different calibration parameters through optimization. Their approach results in little difference from the initial calibration values in terms of intrinsic parameters. Instead, they focus on the use of their technique for estimating line delay in rolling shutter cameras and for inferring exposure time from motion blur. Rehder \emph{et al.} select pixels for rendering where they find large image gradients in the calibration image. Our pixel selection scheme is different from theirs: we use all the pixels that the target is projected to, and our objective function is different.

In more recent work, Rehder and Siegwart\cite{rehder2017camera} extend their direct formulation of camera calibration\cite{rehder2017direct} to include calibration of inertial measurement units (IMUs). In this work, the authors introduce blurring into their renderings to simulate imperfect focusing and motion blur. We also use blurring, and their objective function is more similar to ours in this work. However, they still only select a subset of pixels for rendering based on image gradients, and they, again, did well in estimating exposure time from motion blur but did not otherwise improve results over the baseline approach.

In terms of improved image features, Ha \emph{et al.}\cite{ha2017deltille} proposed replacing the traditional checkerboard with a triangular tiling of the plane (a deltille grid). They describe a method for detecting this pattern and checkerboards in an image and introduce a method for computing the sub-pixel location of corner points for deltille grids or checkerboards. This is based on resampling of pixel intensities around a saddle point and fitting a polynomial surface to these. We consider this approach state-of-the-art in camera calibration based on detection of interest points, and we therefore use it for performance comparison.

\section{METHOD}

\begin{figure}[t]
\captionsetup[subfigure]{justification=centering}
    \centering
    \tikzstyle{decision} = [diamond, draw, fill=gray!10, 
    text width=4em, text badly centered, node distance=3cm, inner sep=0pt]
\tikzstyle{block} = [rectangle, draw, fill=gray!10, 
    text width=3cm, text centered, rounded corners, minimum height=6cm]
\tikzstyle{line} = [draw, -latex']
\tikzstyle{cloud} = [draw, ellipse,fill=gray!10, node distance=3cm,
    minimum height=2em]
    \resizebox{\textwidth}{!}{
    \begin{tikzpicture}[node distance = .4cm, auto]
    \node [block] (init) {Compute \mbox{initial} camera \mbox{calibration} with \mbox{existing} method.
    \subfloat[Real image with detected corners]{\includegraphics[width=\textwidth]{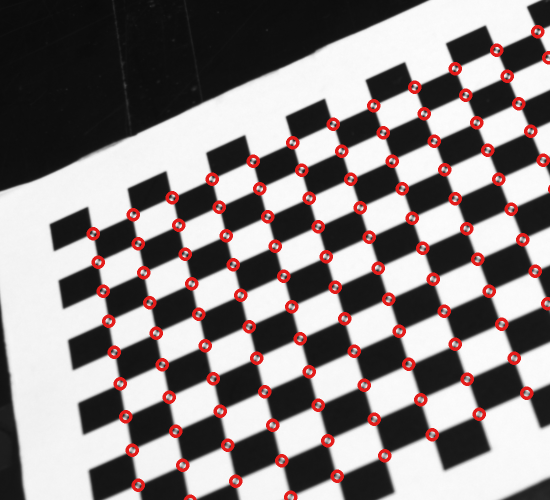}}};
    \node [block, right =of init] (identify) {Render images of calibration objects using estimated parameters.
    \subfloat[\newline Rendered image]{\includegraphics[width=\textwidth]{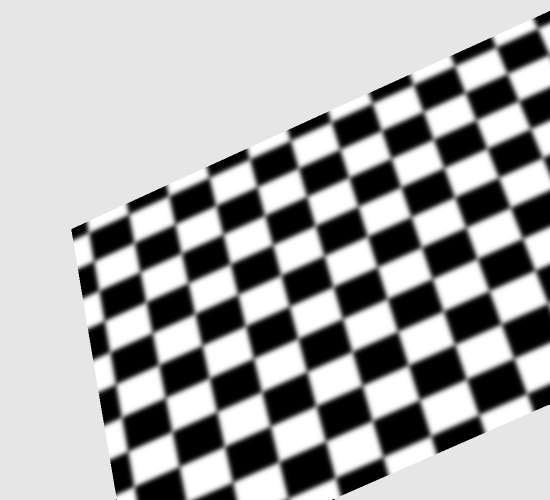}}};
    \node [block, right =of identify] (evaluate) {Update parameters to minimize sum of squared differences. \ \\  \ \\
    \subfloat[Difference between image and rendering]{\includegraphics[width=\textwidth]{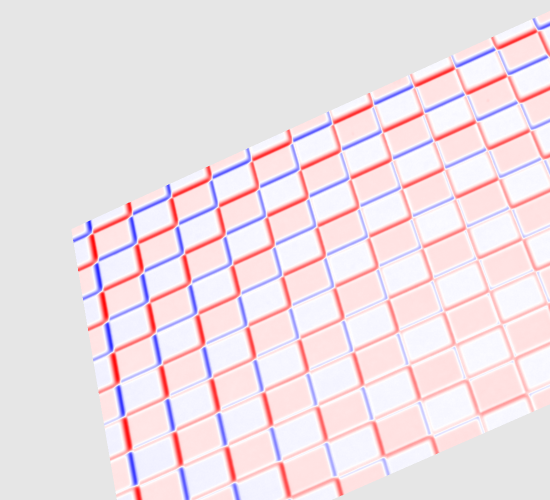}}};
    \node [cloud, below of=evaluate, node distance=3.7cm] (update) {Next iteration};
    \node [decision, right of = evaluate, node distance =2.9cm] (decide) {\scriptsize Converged};
    \node [block, right of=decide, node distance =3cm, minimum height=1cm] (stop) {{\small Difference between image and rendering for \mbox{converged} parameters}\\{\includegraphics[width=\textwidth]{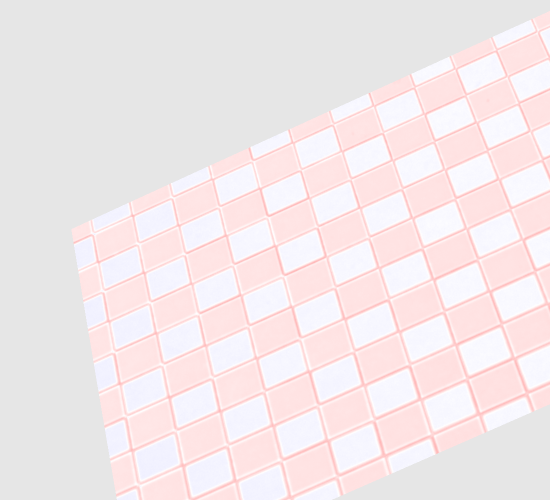}}};
    \path [line] (init) -- (identify);
    \path [line] (identify) -- (evaluate);
    \path [line] (evaluate) -- (decide);
    \path [line] (decide) |- node [near start] {no} (update);
    \path [line] (update) -| (identify);
    \path [line] (decide) -- node {yes}(stop);
\end{tikzpicture}}
\vspace{-6pt}
    \caption{Overview of our method with a checkerboard as an example. All images are crops of a larger image. Difference images: Red represents positive values and blue represents negative values. For clear visualization, we have multiplied the focal length by 1.01 in our initial guess. Note that the converged difference image still resembles a checkerboard pattern because we do not compensate for the board's albedo.}
    \label{fig:overview}
\end{figure}
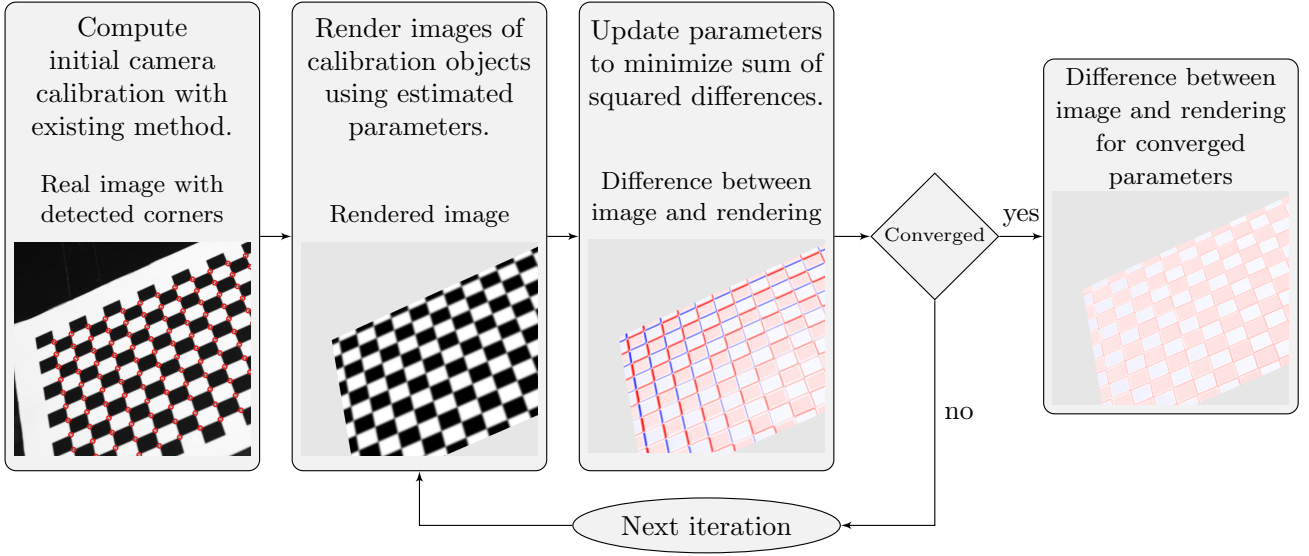

Our method builds on top of an existing camera calibration method. This is used
as a starting guess for the camera matrix $\mathbf{K}_0$, the distortion coefficients $\mathbf{d}_0$ and the poses of each calibration object $\mathbf{R}_{i0}$, $\mathbf{t}_{i0}$. We use this to render images of calibration objects, which we compare with images captured by the camera. Based on this comparison, the optimizer updates the camera calibration until the result is satisfactory. An outline of our method is in \cref{fig:overview}.

The rendering is based on sampling a smooth function $\mathrm{T}_\mathrm{G}$ that describes the texture of the calibration object. The initial calibration with a standard technique reveals which pixels the calibration target covers. Each of these pixels is undistorted and projected onto the surface of the calibration object to get the coordinates for where to sample $\mathrm{T}_\mathrm{G}$. This inverse mapping from pixel coordinates to calibration object coordinates is advantageous with respect to calculating the gradient used in optimization. Each sampled $\mathrm{T}_\mathrm{G}$ value is compared with the intensity of its corresponding pixel, and the sum of squared errors is the objective function that we minimize.

\subsection{Projection of points}
\newcommand{\ijpoint}{\ensuremath\begin{bmatrix}x\\y\end{bmatrix}}
\newcommand{\pp}{\ensuremath\mathrm{P}} 
Let us introduce a function that projects points in $\mathbb{R}^3$ to the image plane of a camera, including distortion
\begin{alignat}{1}
    \pp\left(\mathbf{K}, \mathbf{d},
    \mathbf{p}\right) = \begin{bmatrix} x \\y\end{bmatrix}.
\end{alignat}
Here, $\mathbf{K}$ is a camera matrix, $\mathbf{d}$ is a vector of distortion coefficients, and $\mathbf{p}$ is the point we are projecting, where the elements are
\begin{equation}
    \mathbf{K} = \begin{bmatrix}f_x&0&x_0\\0&f_y&y_0\\0&0&1\end{bmatrix}, \quad 
    \mathbf{d} = \begin{bmatrix}k_1&k_2&p_1&p_2 \end{bmatrix}, \quad 
    \mathbf{p} = \begin{bmatrix}p_x\\p_y\\p_z\end{bmatrix}.
\end{equation}
$\pp$ is then implemented as follows. First, the points are projected to normalized image coordinates:
\begin{equation} \label{eq:project_normalized}
    \begin{bmatrix} x_\mathrm{n}\\y_\mathrm{n} \end{bmatrix}
    =\frac{1}{p_z} \begin{bmatrix}p_x\\p_y\end{bmatrix}, \quad  r^2 = x_\mathrm{n}^2+y_\mathrm{n}^2.
\end{equation}
The normalized points are distorted using the distortion model of Brown-Conrady\cite{brown1966decentering,conrady1919decentred}.
\begin{equation}\label{eq:distortion}
    \begin{bmatrix}x_\mathrm{nd}\\y_\mathrm{nd} \end{bmatrix} = 
    \begin{bmatrix}x_\mathrm{n}\\y_\mathrm{n} \end{bmatrix}\left(1+ k_1r^2+ k_2r^4\right) + 
    \begin{bmatrix} 2p_1x_\mathrm{n}y_\mathrm{n}+p_2\left(r^2+2x_\mathrm{n}^2\right)\\ 2p_2x_\mathrm{n}y_\mathrm{n}+p_1\left(r^2+2y_\mathrm{n}^2\right) \end{bmatrix},
\end{equation}
and the distorted points are converted to pixel coordinates 
\begin{equation}\label{eq:un-normalize}
    \begin{bmatrix}x\\y\end{bmatrix} = \begin{bmatrix}f_x x_\mathrm{nd} + x_0\\f_y y_\mathrm{nd} + y_0 \end{bmatrix}.
\end{equation}
\subsection{Rendering}
Let each calibration board have its own $u,v$ coordinate system, and let $\mathbf{R}_i$ and $\mathbf{t}_i$ describe the pose of the $i$th board. Let $\mathbf{r}_{i_j}$ denote the $i$th column of $\mathbf{R}_i$.
The 3D position of a point on a board is then
\newcommand{\rtboardrow}[1]{R_{i_{#1,1}} & R_{i_{#1,2}} & t_{i_#1}}
\begin{alignat}{1} \label{eq:prt}
    \mathbf{p}(i,u,v) &= \begin{bmatrix}\mathbf{R}_i & \mathbf{t}_i\end{bmatrix} \begin{bmatrix} u\\v\\0\\1 \end{bmatrix}
                     =\begin{bmatrix}\mathbf{r}_{i_1} &\mathbf{r}_{i_2} & \mathbf{t}_i \end{bmatrix} \begin{bmatrix} u\\v\\1 \end{bmatrix}.
\end{alignat}
Using a camera matrix  $\mathbf{K}$ and distortion coefficients $\mathbf{d}$, we can project this point to the image plane
\begin{alignat}{2}
    \ijpoint &= \pp(\mathbf{K}, \mathbf{d}, \mathbf{p}(i,u,v)).
\end{alignat}
We can solve for $u,v$ in terms of $x,y$ in the above expression and obtain a new function:
\begin{equation}
\label{eq:pinv}
    \begin{bmatrix}u\\v\end{bmatrix} = \pp^{-1}(\mathbf{K}, \mathbf{d}, i, x, y).
\end{equation}
As mentioned, we use a function to define the texture of the calibration board (checkerboard, circles, deltille grid, or similar). Let us call this texture function $\mathrm{T}(\mathbf{p}_{uv})$. Because natural images are not always sharp, we introduce a blurry version of the texture map by convolving it with a Gaussian kernel in $u$ and $v$. This has the additional advantage that the texture map becomes smooth, which makes the objective function differentiable.
\begin{equation}
\label{eq:tg}
    \mathrm{T}_\mathrm{G}(\mathbf{p}_{uv}, \sigma_u, \sigma_v) = (G_{\sigma_u} \ast G_{\sigma_v} \ast \mathrm{T})(\mathbf{p}_{uv}).
\end{equation}
This blur is applied in texture space, but we actually want it to be uniform around the interest point in image space. Thus, the standard deviations $\sigma_u$ and $\sigma_v$, need to be corrected according to the length of the positional differential vector of the projection to the image plane. We introduce this quantity as
\begin{equation}\label{eq:diff_norm}
    M_u = \norm{\frac{\partial\pp(\mathbf{K}, \mathbf{d}, \mathbf{p}(i,u,v))}{\partial u}}_2 .
\end{equation}
Inserting \cref{eq:pinv,eq:diff_norm} in \cref{eq:tg}, we obtain a function that enables the rendering of an image of the calibration object:
\newcommand{\render}{\mathrm{C}_i}
\begin{equation}\label{eq:render}
    \render(x,y, \sigma_{i,j}) = \mathrm{T}_\mathrm{G}\left(\pp^{-1}\left(\mathbf{K}, \mathbf{d}, i, x, y\right),  
    \sigma_{i,j} / M_u,
    \sigma_{i,j} / M_v \right),
\end{equation}
where $M_v$ is the same as $M_u$ but with respect to $v$ and $\sigma_{i,j}$ is a measure of how blurry the image is around the $j$th interest point on the $i$th calibration board. This implies that the formula is only valid in the neighborhood of this point, and therefore we introduce
\begin{equation}\label{eq:nij}\mathcal{N}_{i,j}\end{equation} 
to describe the set of pixel coordinates where the rendering is accurate. We choose $\mathcal{N}_{i,j}$ to be the pixels where the corresponding $u,v$ coordinate is no further away than one half of the interest point spacing in Manhattan distance given by the initial camera calibration.
For convenience of notation, let us define a set containing all $\mathbf{R}_i$, $\mathbf{t}_i$, and $\sigma_{i,j}$
\begin{equation}\label{eq:beta}
    \beta = \big\{\mathbf{R}_i, \mathbf{t}_i : i\in \{1, \dots, n_i\}\big\}\cup \big\{\sigma_{i,j} : i\in \{1, \dots, n_i\}, j\in \{1,\dots, n_j\}\big\},
\end{equation}
where $n_i$ is the number of calibration boards and $n_j$ is the number of interest points on each calibration board.
Using \cref{eq:render,eq:beta,eq:nij}, our optimization problem is then
\begin{equation}\label{eq:objfun}
    \hat{\mathbf{K}}, \hat{\mathbf{d}}, \hat{\beta} = \argmin\limits_{\mathbf{K}, \mathbf{d}, \beta} 
    \sum\limits_{i=1}^{\smash{n_i}} \sum\limits_{j=1}^{\smash{n_j}} \sum\limits_{x,y\in \mathcal{N}_{i,j}} 
    \Big(\render(x,y, \sigma_{i,j})- I_i(x,y)\Big)^2,
\end{equation}
where $I_i(x,y)$ is the intensity of the pixel at $x,y$ in the image containing the $i$th calibration board.
We parameterize $\mathbf{R}_i$ as quaternions and solve \cref{eq:objfun} using the Levenberg-Marquardt algorithm\cite{levenberg1944method,marquardt1963algorithm}.
Because \cref{eq:tg} is defined to give values between 0 and 1, in the case where $\sigma=0$, our optimization problem is equivalent to maximizing the sum of pixels on white parts of $\mathrm{T}$ while minimizing the sum of pixels corresponding to black parts of $\mathrm{T}$.

\subsection{Computation of $\mathrm{P}^{-1}$}
Recall that $\mathrm{P}^{-1}$ is the function that, given a camera calibration and the pose of a calibration board, transforms from $x,y$ in pixel space to $u,v$ coordinates on the board. The first step in computing this is to invert \cref{eq:un-normalize} by normalizing the pixel coordinates
\begin{equation}
    \begin{bmatrix}x_\mathrm{nd}\\y_\mathrm{nd} \end{bmatrix} = \begin{bmatrix}\frac{x-x_0}{f_x}\\\frac{y-y_0}{f_y} \end{bmatrix} .
\end{equation}
Inverting \cref{eq:distortion} is not possible to do analytically, so we use an iterative numerical approach\cite{opencv_library}. Note however that we can compute analytical derivatives of the inverse of \cref{eq:distortion} by applying the inverse function theorem.
To map the undistorted normalized coordinates to the calibration object, we combine \cref{eq:prt,eq:project_normalized}:
\begin{equation}
    s\begin{bmatrix} x_\mathrm{n}\\y_\mathrm{n}\\1 \end{bmatrix}  = \underbrace{\begin{bmatrix}\mathbf{r}_{i_1} &\mathbf{r}_{i_2} & \mathbf{t}_i \end{bmatrix}}_{\mathbf{H}_i} \begin{bmatrix}u\\v\\1 \end{bmatrix}.
\end{equation}
From this, it is clear that $\mathbf{H}_i$ is a homography transforming from the space of the $i$th calibration board to the normalized image plane. We invert the homography to perform the mapping
\begin{equation}
    s\begin{bmatrix} u\\v\\1 \end{bmatrix}  = \mathbf{H}_i^{-1} \begin{bmatrix}x_\mathrm{n}\\y_\mathrm{n}\\1 \end{bmatrix}.
\end{equation}
Because the Levenberg-Marquardt algorithm is gradient-based, we need derivatives. We designed our texture function $\mathrm{T}_\mathrm{G}$ to be smooth and differentiable, and fortunately the function $\pp^{-1}$ is also differentiable, which implies that $\render$ is differentiable. Our implementation uses dual numbers for computing analytical derivatives.

\section{RESULTS}
When comparing a camera calibration to the ground truth, one could measure errors of each parameter individually,\cite{zhang2011camera} but this is difficult to interpret, especially for distortion parameters as they can counteract each other. Motivated by this, we introduce \emph{per-pixel reprojection error}, which measures the root mean squared distance in pixels between points projected with the true and estimated camera intrinsics. For each pixel, the image plane coordinates $x,y$ define a line in $\mathbb{R}^3$ along which we select a point $\mathbf{q}_{xy}$ that projects to this pixel:
\begin{equation}
    \pp(\mathbf{K}, \mathbf{d}, \mathbf{q}_{xy}) = \ijpoint,
\end{equation}
where $\mathbf{K}$ is the true camera matrix and $\mathbf{d}$ are the true distortion coefficients. We can now compute the per-pixel reprojection error $E$ by using the estimated parameters to project the same points. Computing the differences, we have
\begin{equation}
    E = \sqrt{\sum\limits_{x,y} \norm{\pp(\hat{\mathbf{K}}, \hat{\mathbf{d}}, \mathbf{q}_{xy})-\ijpoint}_2^2},
\end{equation}
where $\hat{\mathbf{K}}$ is the estimated camera matrix, $\hat{\mathbf{d}}$ are the estimated distortion coefficients and $x,y$ sum over all possible pixel locations. 

\begin{figure}[t]
    \centering
    \includegraphics[width=\textwidth]{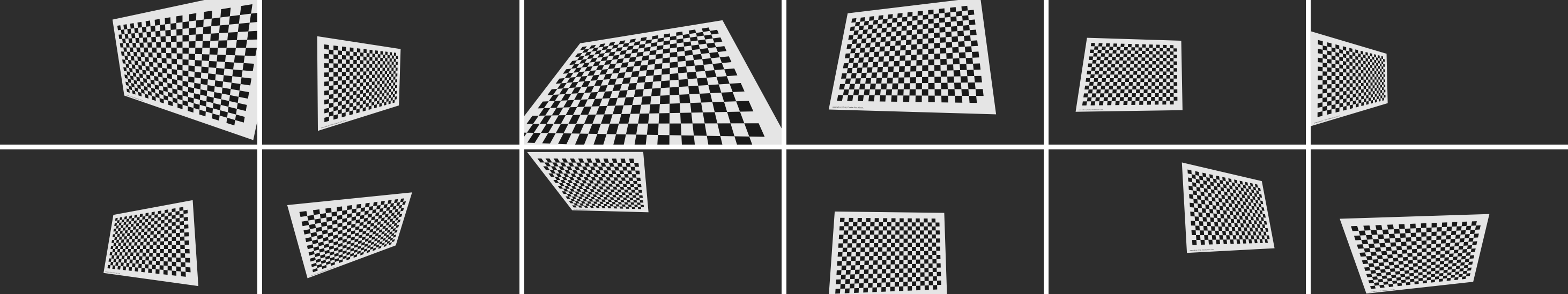}
    \\[1ex]
    \caption{Sample images from our synthetic image dataset.}
    \label{fig:data_synth}
\end{figure}
\begin{figure}[t]
    \centering
    \subfloat[$\sigma=0.5$, $\sigma_n = 0$]{\includegraphics[width=.15\textwidth]{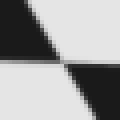}}\
    \subfloat[$\sigma=0.5$, $\sigma_n=1.5$]{\includegraphics[width=.15\textwidth]{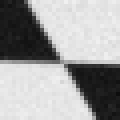}}\
    \subfloat[$\sigma=0.5$, $\sigma_n = 3$]{\includegraphics[width=.15\textwidth]{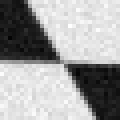}}\
    \quad
    \subfloat[$\sigma=0$, $\sigma_n = 1$]{\includegraphics[width=.15\textwidth]{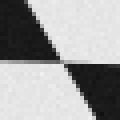}}\
    \subfloat[$\sigma=1$, $\sigma_n = 1$]{\includegraphics[width=.15\textwidth]{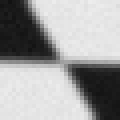}}\
    \subfloat[$\sigma=2$, $\sigma_n = 1$]{\includegraphics[width=.15\textwidth]{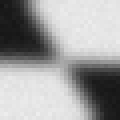}}
    \vspace{4pt}
    \caption{A corner from a checkerboard in a synthetic image with various levels of blur and noise added.}
    \label{fig:blur_noise}
\end{figure}

\subsection{Synthetic data}
We generate a set of 500 images of size $1920\times 1080$ with a virtual camera with focal length $f=1000$ each containing a single $17\times24$ checkerboard. The images are rendered so that the pixel intensities lie in the range $[0.1;0.9]$. \Cref{fig:data_synth} shows examples of these images. Each image is blurred by filtering it with a Gaussian kernel with zero mean and standard deviation $\sigma$. After this we add normally distributed noise to each pixel with zero mean and standard deviation $\sigma_n$, examples of this can be seen in \cref{fig:blur_noise}.

\begin{figure}[h]
    \centering
    \includegraphics[width=8cm]{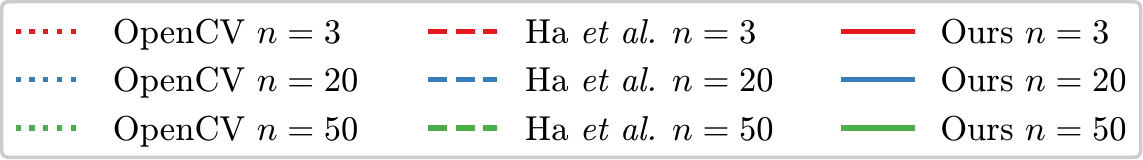}\\
    \includegraphics[width=.49\textwidth]{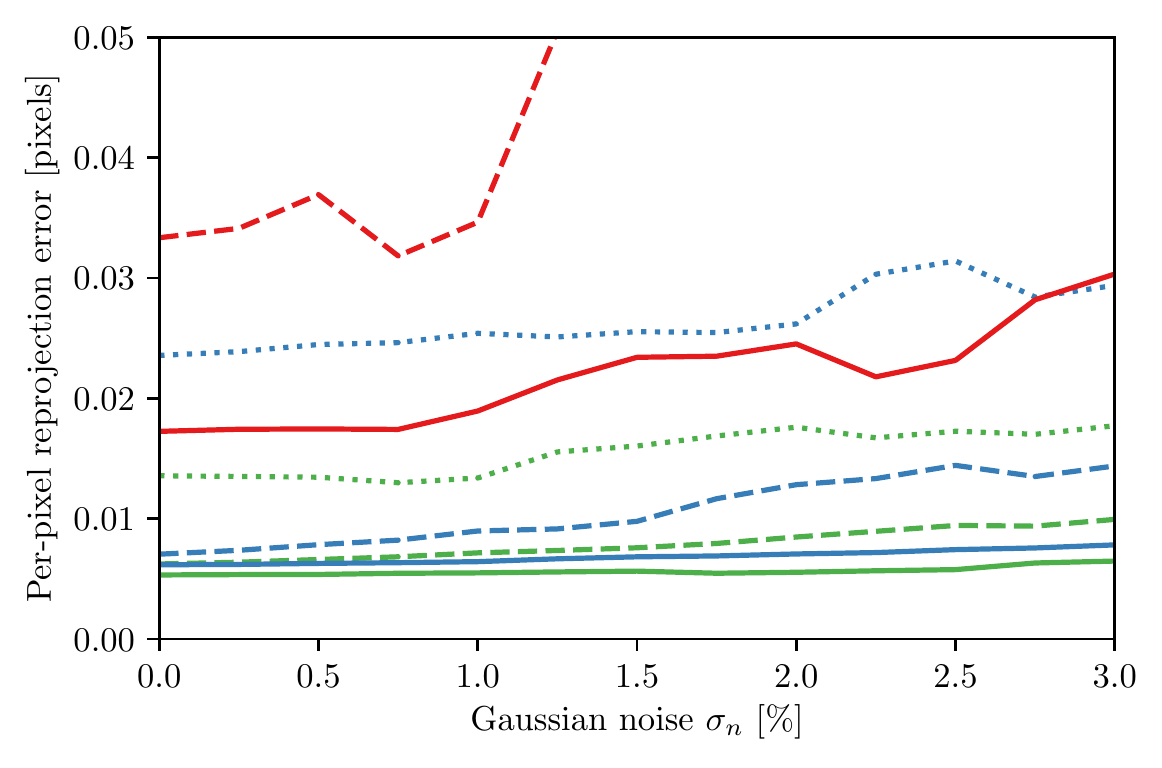}
    \includegraphics[width=.49\textwidth]{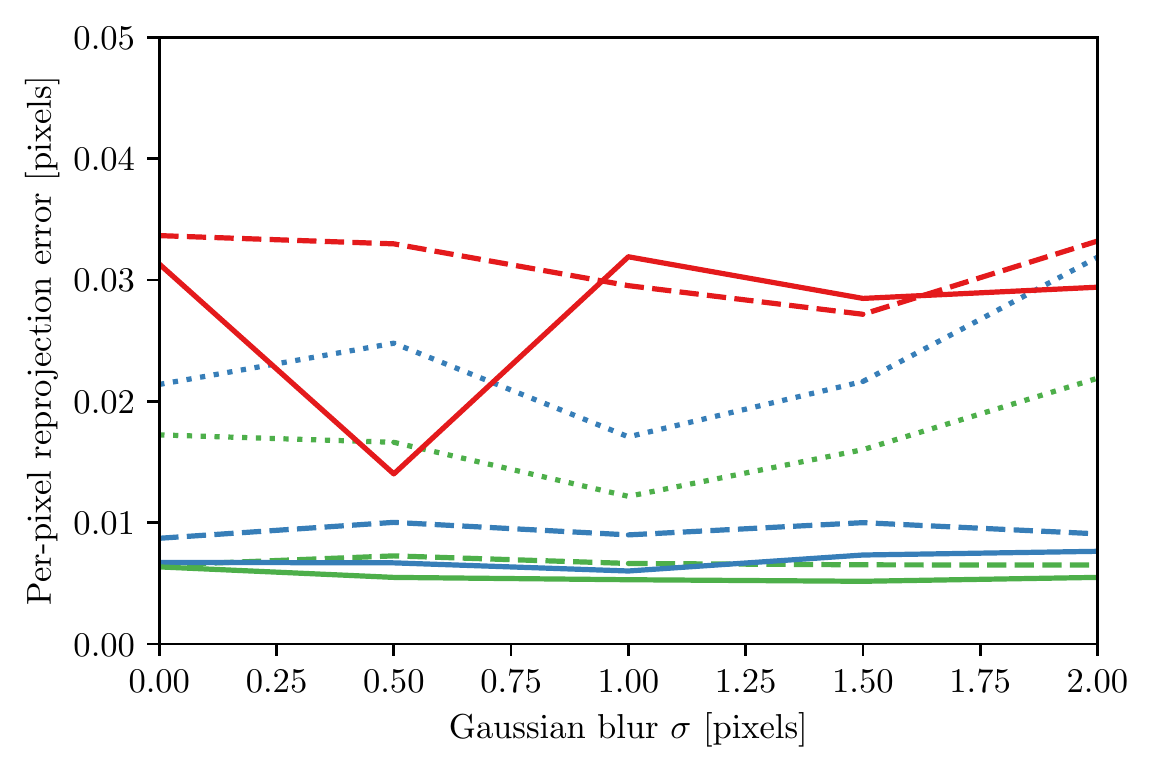}
    \caption{Comparison of our method with Ha \emph{et al.}\cite{ha2017deltille} and OpenCV\cite{forstner1987fast,opencv_library} for varying number of images used in calibration ($n$). Left: Varying $\sigma_n$ with fixed $\sigma=0.5$. Right: Varying $\sigma$ with fixed $\sigma_n=1\%$. OpenCV $n=3$ lies beyond the plotted area.}
    \label{fig:synth_graph}
\end{figure}

We select $n$ random images from these and use the checkerboard detector from Ha \emph{et al.}\cite{ha2017deltille} with default parameters to detect points. After this, we use the standard method by Zhang\cite{zhang2000flexible} to compute the camera calibration. This is a calibration we compare with (Ha \emph{et al.}), but also our initial guess for \cref{eq:objfun}. We do this for $n\in\{3, 20, 50\}$ and for varying values of $\sigma$ and $\sigma_n$. For each $n$, $\sigma$ and $\sigma_n$ we perform 25 trials with randomly sampled images. The results of these experiments are in \cref{fig:synth_graph}. For comparison with OpenCV\cite{opencv_library}, we use the detected points as initialization for \texttt{cornerSubPix}\cite{forstner1987fast}, with a $5\times5$ window. As the images are rendered without distortion, we do the calibration without distortion as well.

We observe that our method performs better than Ha \emph{et al.}\cite{ha2017deltille} and OpenCV\cite{forstner1987fast,opencv_library} for each $n$ across various levels of noise and blur, except for $n=3$ in cases with much blur. We also observe that our method is consistently better in noisy situations.

\begin{figure}[h]
    \centering
    \includegraphics[width=\textwidth]{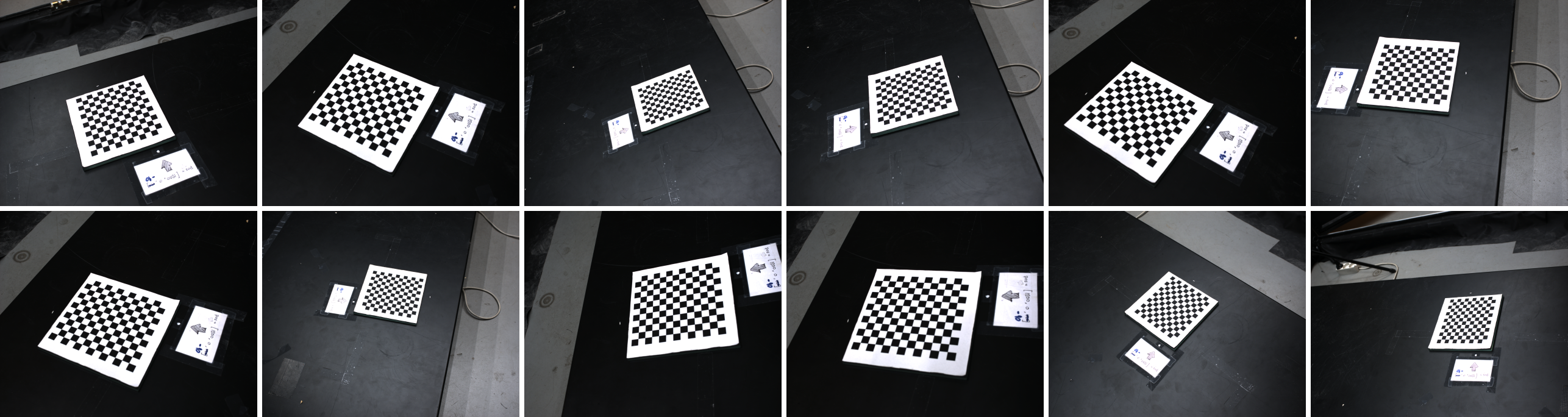} \\[1ex]
    \caption{Sample images from our real image dataset.}
    \label{fig:data_real}
\end{figure}

\subsection{Real data}
When comparing camera calibrations on real data, an often reported measure is the reprojection error of all points. That is however not something we are able to do as our method incorporates the constraint of the calibration object geometry, and the reprojection error will thus per definition always be zero. Based on \cref{fig:synth_graph}, the points detected by the detector from Ha \emph{et al.}\cite{ha2017deltille} are clearly quite accurate, which motivates us to use it as a pseudo-ground truth. 

We use a dataset of 128 images at a resolution of $3376\times2704$, each containing a $12\times13$ checkerboard. We randomly select 64 images to use as our test set, and detect points in them with Ha \emph{et al.}\cite{ha2017deltille} which we use as pseudo-ground truth. Then we select $n$ of the images not in the test set and use them to compute the camera intrinsics. For each image in the test set, we use the already detected points together with our camera calibration, to compute the pose of the checkerboard, which in turn allows us to project points to the camera, and thereby measure a reprojection error. 
\begin{figure}[b]
    \centering
    \includegraphics[width=.7\textwidth]{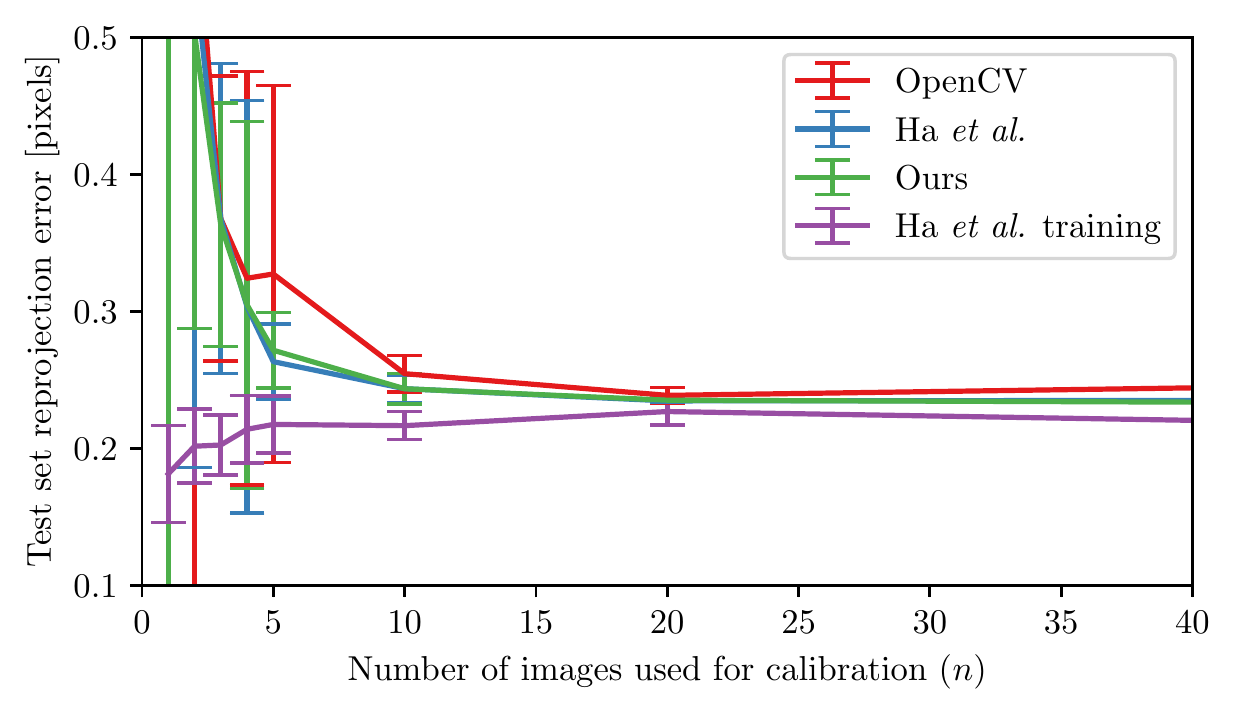}
    \caption{Reprojection error for images in the test set of our real dataset as a function of $n$. Bars on each point show $\pm1$ standard deviation.}
    \label{fig:real_graph}
\end{figure}
For each $n$ we partition the 64 images in the training set into non-overlapping sets of size $n$ and do the camera calibration for each of them. \Cref{fig:real_graph,tab:real_numbers} show the performance of our method compared to Ha \emph{et al.}\cite{ha2017deltille} and OpenCV\cite{forstner1987fast}. For $n<5$ our method performs better and has a lower standard deviation. For large $n$, Ha \emph{et al.}\cite{ha2017deltille} achieve an extremely similar reprojection error, but the points we are computing the reprojection error against are also detected by their method. It can also be seen that the reprojection error of their method on the training data approaches the same values from below, so the best achievable test set reprojection error is limited either by the camera model or the accuracy of the detected points.

\begin{table}[h]
    \centering
    \caption{Data from \cref{fig:real_graph}. $\pm$ indicates the standard deviation of the reprojection error.}
    \vspace{4pt}
    \label{tab:real_numbers}
    \begin{tabular}{lccccc}
    \toprule
    $n$ & 2 & 3 & 4 & 5\\
\midrule
OpenCV&$0.61\pm0.52$&$0.37\pm0.10$&$0.32\pm0.15$&$0.33\pm0.14$\\
Ha \emph{et al.}&$0.55\pm0.36$&$0.37\pm0.11$&$0.30\pm0.15$&$0.26\pm0.03$\\
Ours&$0.50\pm0.22$&$0.36\pm0.09$&$0.30\pm0.13$&$0.27\pm0.03$\\
\bottomrule
    \end{tabular}
\end{table}

\section{DISCUSSION}
Although we have only used this method to compute intrinsics of a single camera in this paper, it is straight-forward to extend to intrinsics of multiple cameras and their extrinsics. The homography $\mathbf{H}_i$ can easily incorporate the pose of the camera, and then all one needs is a separate set of parameters per camera.

Even though we have chosen to use the Brown-Conrady\cite{brown1966decentering,conrady1919decentred} distortion model in our work, this is a choice mostly motivated by being able to fairly compare with OpenCV\cite{opencv_library}. Our method is not tailored to this distortion model, and one could replace it with another, such as the division model\cite{fitzgibbon2001simultaneous}.

We do not attempt to match the scaling of the image intensities in the rendering as in the work of Rehder \emph{et al.}\cite{rehder2017direct} We experimented with scaling the image intensities to match the rendering or including the local intensity of the rendering as a parameter in the optimization as well, but we did not observe any increase in accuracy when doing this.

Our method takes around three minutes to solve the optimization problem for 40 images from our real dataset, where each image is 9 Megapixels. We find this to be an acceptable computation time, especially given that even one such problem contains around 30 million residuals.

\section{CONCLUSION}
We have introduced a method for improving camera calibration based on minimizing the sum of squared differences between real and rendered images of textured flat calibration objects. 
Our rendering pipeline consists purely of analytically differentiable functions, which allows for exact gradients to be computed making the convergence of the optimization more robust and fast, while still allowing us to blur the image in the image space as would naturally occur.
On synthetic data, our method outperforms state-of-the-art camera calibration based on point detection, for images distorted by Gaussian blur and noise. 

On real data, our method exhibits a clear advantage when only a few images are available for calibration, and performs at least as well for a larger number of images, but we have not been able to verify whether our method outperforms the existing methods in this case, due to the difficulty of evaluating which of two estimated camera intrinsics is better.

\bibliography{report} 
\bibliographystyle{spiebib} 

\end{document}